# Performance analysis of facial recognition: A critical review through glass factor

Jiashu HE
Dept. Computer Science
*University of Illinois, Urbana Champaign*
Champaign, US
jiashuh2@illinois.edu

*Abstract*— COVID-19 pandemic and social distancing urge a reliable human face recognition system in different abnormal situations. However, there is no research which studies the influence of glass factor in facial recognition system. This paper provides a comprehensive review of glass factor. The study contains two steps: data collection and accuracy test. Data collection includes collecting human face images through different situations, such as clear glasses, glass with water and glass with mist. Based on the collected data, an existing state-of-the-art face detection and recognition system built upon MTCNN and Inception V1 deep nets is tested for further analysis. Experimental data supports that 1) the system is robust for classification when comparing real-time images and 2) it fails at determining if two images are of same person by comparing real-time disturbed image with the frontal ones.

## I. FACIAL RECOGNITION SYSTEM

### A. Background and related work

With the ever-accelerated update of science and technology, all kinds of computers and digital devices have changed people's lives fundamentally. The merge of online payment systems, e-commerce and remote work apps lead to an urgent need for a safer and more efficient authentication method than using password. Aside from unlocking the smart phone, facial recognition technology made a remarkable contribution to anti cyber-attacks, terrorists, identifying children and supporting investigations.

Despite significant progress in this field, implementing facial identification efficiently and robustly has presented serious challenges. For instance, the Thatcher effect and Double illusion can heavily influence the accuracy of a facial recognition technique [1]. There exist tailor-made models for some of these problems under special circumstances. However, none of these approaches are robust enough for unconstrained facial recognition problems in general. Among the numerous techniques applied to settle the facial recognition problem, several models and mathematical approaches serve as the major building blocks for facial recognition applications, which include eigenfaces, neutral networks and graph matching.

Eigenface aims to extract the specific facial information by reconstructing the image in the eigen space [2]. The eigen vectors of the covariance matrix are ordered according to the corresponding eigen values to represent different amounts of the variation. Each input image can be represented as one and only one combination of the eigen vectors. One intuitive way to determine the corresponding class the image belongs to is to minimize the Euclidean distance between each face class in the Euclidean space. One outstanding feature of eigenface is its simple and efficient approach of face matching. However, this method deals with the input face as a holistic image, while not considering any specific aspects such as illumination or pose.

Neural networks have arguably been the best-performing algorithm used to settle different problems in the field of artificial intelligence [3], such as neutral language processing [4], all kinds of prediction and of course, facial recognition. Neutral networks classify the faces by learning a weight on each node based on the large amount of input data, usually labeled [5]. Modeled by the human brain, the network consists of thousands of processing nodes organized into layers [6]. When an input data is given, the node multiplies the input data with its weight, and pass the result to the activation function. Then the activation function will decide whether a node is still alive or dead for input to the next layer. This process is similar to when human brains judge if the given object has some partial features of an object [7]. As the layers move forward, the feature becomes more and more holistic and finally output a value that represents the likelihood that a given object belongs to a particular class. One of the most attractive parts of using neutral networks is to stack liner functions with a non-linear layer in between. In this way multiple templates can be used for one class in one layer and combine them in a later layer.

The performance of a neutral network model heavily relies on the structure of which the layers of nodes are stacked. For instance, VarGFaceNet [8] is an efficient variable group convolutional network for light weight face recognition. For lightweight network, discriminative ability is required as much as possible, so 3x3 row with stride 1 is applied. The structure also features in the fully connected layer for embedding faces. The diversity of structures is a major reason why this algorithm is hardly robust for the general non-constrained input either. Though the majority of models perform satisfactorily for fontal images, different structures are needed for different sets of problems. For example, faces under specific illusion environment or behind the glasses. What's worse, the "black box" nature makes it very difficult to debug if a prediction is wrong, most of the results come from experiments based on input data of large scale, which makes it extremely challenging to find more accurate or robust structures.

Graph matching is the problem for finding similarity between graphs. The graph nature of photo representation inside computers make it a remarkable method for face recognition problem. Wiskott [9] and his colleagues announced a system utilizing Gabor wavelet transform for recognizing human faces represented by labeled graphs. Features of new face images are extracted by elastic graph matching process. The similarity compare function is simple but computationally expensive, but the use of dynamic link architecture makes it robust to rotation invariance compared to other techniques.

Despite the significant progress stated above in the field of face recognition, the majority of the satisfactory-performing models require high-quality frontal images as input in order to correctly classify the human face. However, during the COVID-19 pandemic, it is common to see people communicate through glasses to comply with social distancing. In India, glass shields are applied to avoid close contact between the customs officers and the travelers. The need to identify people with a glass in between presents serious challenges to current approaches. In the United states, CDC has recommended social distancing as the best practice to slow down the spread of virus [10], in which the none-contact communication relies heavily on glasses as the intermediate for it is both transparent and well seamless.

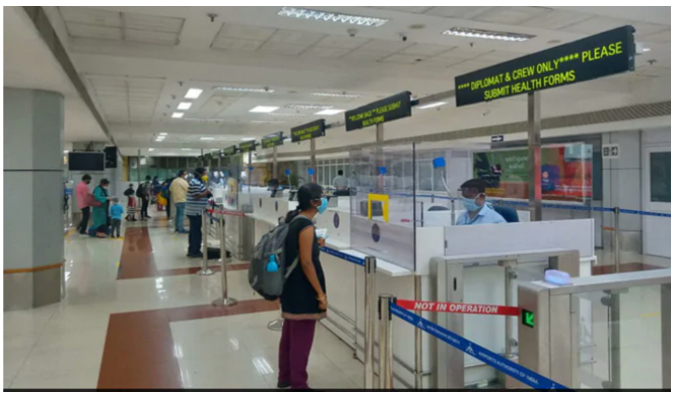

Fig. 1.  At the airport counters, glass shields are used to avoid close contacts[11]

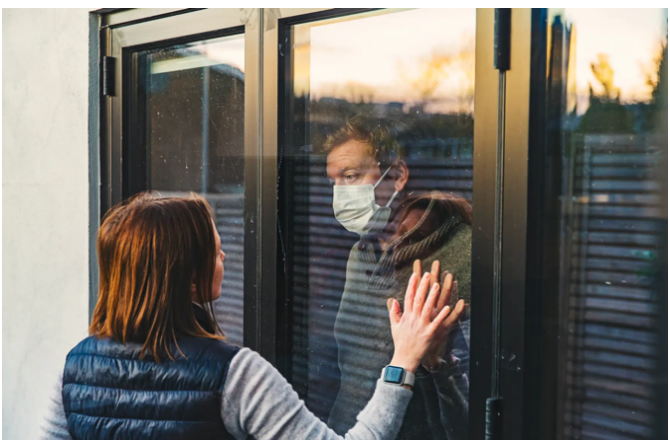

Fig. 2.  Social distancing in the United States[12]

This paper provides a study that evaluates the accuracy of the implementation based on the MTCNN face detection network and Inception V1 face classification network using a dataset with glass and water drop as interference. In section II, the face recognition system to be tested will be introduced together with the deep nets it utilizes for detection and recognition, respectively. Then the data collection process and experiment setup will be provided in section III, together with the observations and related conclusions drawn about the system.

*B. Holistic view of the system*

Face recognition is the task to enable computers to identify human beings from their photographs. Modern recognition systems usually separate the detector and the classifier thus that the cropped faces can be saved for further use. The separation of detection and recognition also makes it easier to debug, thus improve the robustness of the system. The recognition stage is usually more computing resource consuming, which can be finished remotely on the server side to make the work easier on the mobile end. Mobile feasibility makes a positive contribution to the fast progress in the field of face recognition and computer vision.

This paper focuses on testing the performance of a face detection and recognition system with regard to its accuracy in identifying photos taken through glasses, water drops and mists. The system consists of 2 sub-systems, a face detection MTCNN network [13] and an Inception V1 [14] deep net for embedding and classification. The MTCNN facial detection model iterates through the data loader, detect the human faces and associate it with the detection probability. The mtcnn function returns the detected human face. The cropped faces are then used as the input of the Inception V1 deep net, the faces of all same size make the batch processing easy and efficient.

*C. The MTCNN face detection model*

The MTCNN network is able to produce the bounding box, five-point face landmarks and the detection probability simultaneously [13]. The system consists of 3 cascaded stages after resizing the input image firstly. The pyramid of images is fed into 3 ordered networks. The first fully convolutional network produces all candidate facial windows and their bounding boxes. Then all the proposed candidates are fed to another CNN to refine and reject the false and overlapped windows. In stage 3, the network further identifies the face region with 5 facial landmarks.

In stage 1, an image pyramid is created so images in different sizes can be searched for different sizes faces within the input. A 12x12 filter is used to convolve through the image and the resulting patches are processed by P-Net, the bounding box is returned if a candidate face is detected. A stride of 2 is applied to reduce the computational cost. Since most images are large compared to 2 pixels, use of stride will not affect the accuracy of the model.

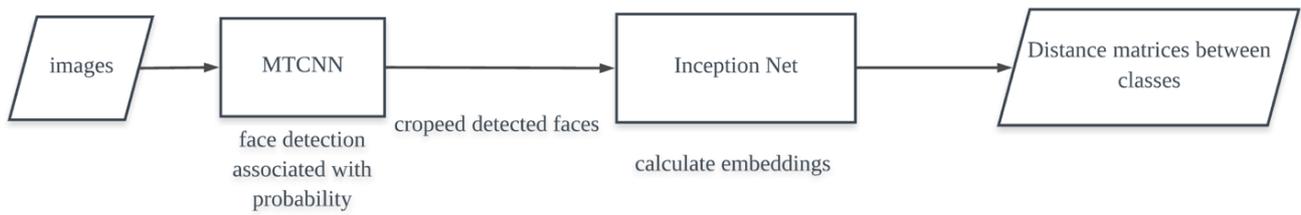

Fig. 3. A face recognition pipeline

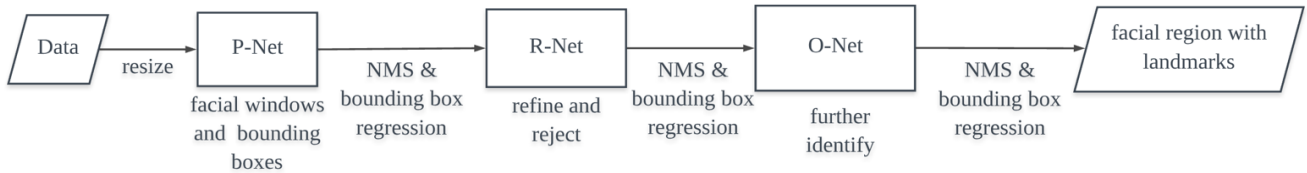

Fig. 4. MTCNN face detection model

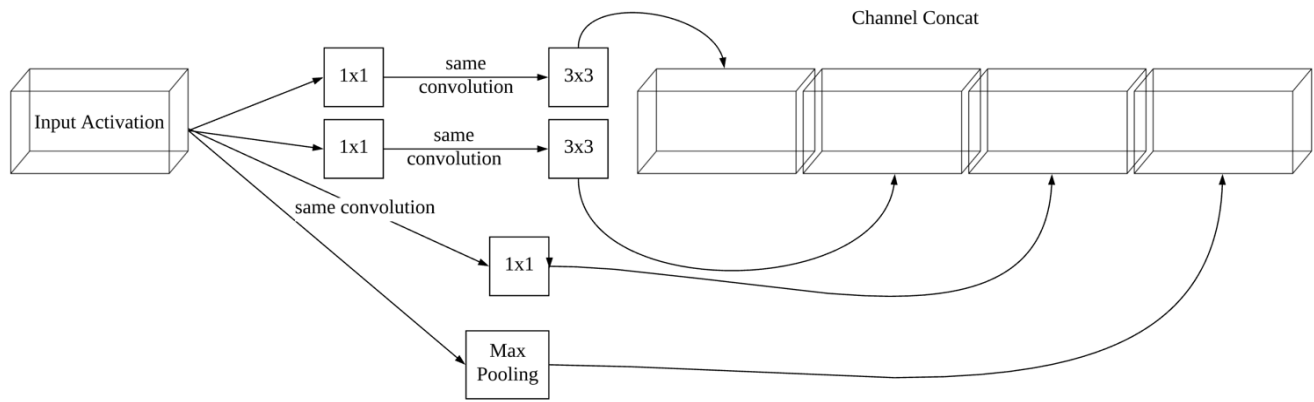

Fig. 5. Inception Layer

P-Net outputs the bounding box result for every 12x12 patch, then the system parses the output to get bounding boxes with higher confidence and standardize the coordinate system by converting all of them to the coordinate system of the un-scaled image. Then Non-maximum Suppression (NMS) method is conducted to further reduce the number of bounding boxes by deleting the boxes that overlaps a lot with the box of highest confidence. The greatly reduces the candidate bounding boxes to be processed, saves a lot of computation for later search. Last but not least, the bounding boxes are squared by expanding the shorter side and fed into the second stage II for further search. To further reduce the number of candidates bounding boxes, the output of stage 1 is converted to 24x24 arrays which are then fed to the R-Net. R-Net outputs the more precise bounding boxes with confidence level.

The O-Net takes in 48x48 bounding boxes from R-Net. Besides the bounding box and the corresponding confidence level, it also outputs the facial landmark coordinates. Similar to the previous 2 levels, the bounding box and facial landmarks with low confidence are dropped. Standardization is applied before output.

1.4 Inception V1 network

Inception net is a kind of networks featuring in inception layer, which achieves a balanced point between the sparsity architecture for better performance and the cost of computational recourses [14]. Instead of considering the size of the filters of the convolutional layer and their tradeoff between the pooling layer. The inception layer tries all of them together and catenate them together as the input of the next layer. To fix the infeasible cost of calculating the activation maps, 1x1 same convolution are applied between the input and the filters.

Inception net is a milestone in the development of convolutional neutral network because the previous models just try to go deeper to achieve better performance. The model is highly creative both in pushing performance and reducing cost. The study [14] argues that when creating a subsequent layer in a deep learning model, one should pay attention to the learnings of the previous layer. To take full usage of the previous layer learning outcome, an appropriate filter size must be achieved [15], the inception model allows the network to choose from different filters and decide which one is the most relevant to learn the required information. The flexibility enables the network to learn for feature abstraction more accurately than the simple structures thus result in higher accuracy in computer vision [16]. The involvement of 1x1 convolution dramatically reduce the computational cost. Either of these features make it a remarkable progress in the development of deep learning models.

## II. GLASS FACTOR STUDY

In this section, the experiment process is presented to evaluate the accuracy of the aforementioned facial recognition system over 4 sets of face images. The facial data are collected through glasses with different disturbing factors. Our experiment tests out the system in a holistic manner before breaking it up to face detection stage and classification stage. Due to the refraction effect dealt by glasses, water and dust, a decline in recognition accuracy and face detection probability is expected. The rest of this section consists of 2 parts, details about the features of the 4 sets of data, how they are collected, and the experiment results over each of them will be given, respectively.

### A. Data Collection

In addition to faces behind crystal glasses like the check-in counter in airport during COVID-19 pandemic, images to be recognized may also be taken by CCTV through windshield. For instance, when the police force analyzes video for criminals. In rainy days, the glasses could be covered by water drops. If the criminal doesn't bother washing cars, the glass that the face is taken though may be full of dust.

In view of the complexity of the environment factors, and the urgent need for a robust recognition system, 4 datasets are included in the experiment. The 4 sets of images are taken under normal condition, through clear glasses, glasses with water flows (when it rains cats and dogs) and glasses with water drops (mimic light rain), respectively.

The images are taken from 10 people of different age groups, with the youngest 4 years old and 84 as the oldest. A crystal glass is applied to create the first dataset where images are taken through ordinary glasses. To simulate water flows and drops, water is poured or sparked onto the glass while the photo is taken through it. Last but not least, nebulizer is applied to create mist on the glass. To make our result significant, all photos of the same identity are on a control-variable basis, the illumination condition and the distance to the camera is the same for all the image data, each identity is asked to show 3 expressions, smile without showing the teeth, smile and show the teeth, and poker face, under which the 3 photos are taken.

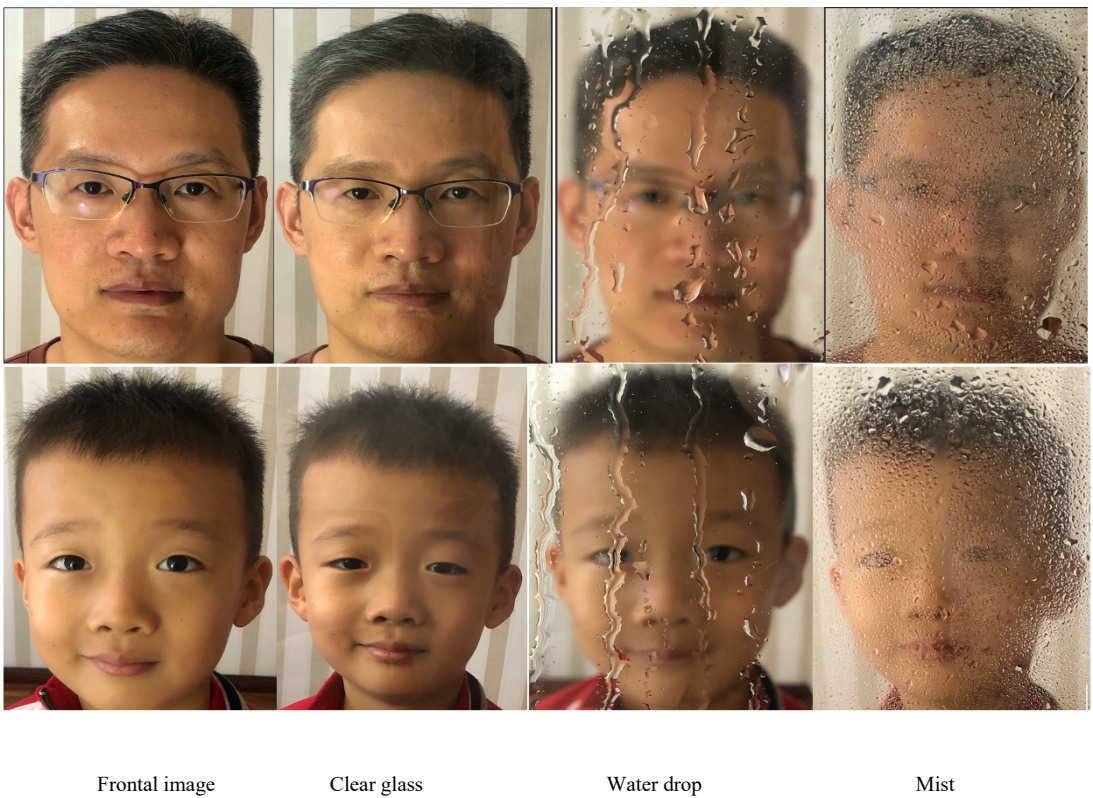

Frontal image      Clear glass      Water drop      Mist

Fig. 6. Images with different disturbing factors (normal frontal image, image taken through clear glass, image taken through glass with water drops, image taken through glass with mist)

## B. Experiment

Face detection and recognition is performed using the pretrained MTCNN and Inception V1 deep net. For each of the identity, 4 sets of data are included in the test, images taken under normal condition, through clear glass, glass with water flows and glass with water drops. Each data set contains 3 different images under the corresponding specific environment factors to be studied. For the total 10 identities, 4 datasets are created for each of them, and 3 images for each one of the datasets, the total number of image data is 120. A horizontal study is performed on these 120 images and output the distance matrix as the evidence for evaluating the model accuracy.

The distance matrix output by the Inception network is of dimension 120 x 120, which is difficult to draw any conclusion from. Photos of a person under a particular environment factor is treated as a unit. For example, the 3 photos of "identity 1" taken through clear glass. For every pair of 2 such units, the average of the 9 distances and take that as the distance between the 2 units. In this way the 120 x 120 matrix is divided to 7 10 x 10 matrixes. The 7 tables correspond to 7 cases, namely normal to normal, normal to water, normal to glass, normal to mist, mist to mist, water to water and glass to glass. In each one of the 7 tables, the 10 rows (columns) correspond to one of the 10 identities (units to be compared). The system is evaluated based on distance changes and classification feasibility.

TABLE I.    DISTANCE CULCULATION

|  | Id1_glass_img1 | Id2_glass_img2 | Id3_glass_img3 |
|---|---|---|---|
| **Id1_glass_img1** | 0 | 0.33 | 0.39 |
| **Id2_glass_img1** | 0.33 | 0 | 0.32 |
| **Id3_glass_img1** | 0.39 | 0.32 | 0 |

a. These 9 distances are averaged as the distance between "1_glass" and "1_glass" in "glass to glass" distance matrix

Of all the 120 images, the MTCNN model performs satisfactorily in extracting faces through clear glasses. However, it failed to detect faces in one image taken through water-filled glass and 2 taken through glass with mist. It fails in 3 cases out of 60 with water drop or mist, resulting a 5% false rate in these 2 conditions. A possible cause for this problem is that the proposal network produces all similar false bounding boxes which are all rejected by the R-net. The model is robust against clear glass in our dataset, but whether it's suitable for applications like detecting through glasses on the airport counter needs to be tested using larger datasets. The pretrained model is able to detect faces through water or mist with a successful rate of 95% in our dataset, better performance is expected if the network is trained on the water or mist images before using it to extract the faces in such applications.

The inception V1 network is tested by comparing the distances it outputs between one identity with itself and other identities. If under the glass, water or mist conditions the distance between one identity and himself increases or between other identities decreases, the model's performance is downgraded. The relative magnitude between the maximum self-compared distance and the minimum inter-identity distance, because that determines whether a threshold can be set to perform classification.

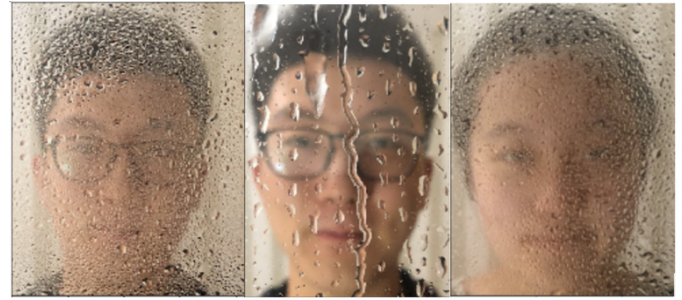

Fig. 7.    Images that MTCNN fails to extract faces from

### 1) Image taken through glass

The clear glass factor has an influence on its overall performance. By comparing the "glass to glass" matrix and "normal to normal" matrix, the average distance between each identity and himself increases by 14.85% and decreases by 2.08%. This comparison is significant for applications where 2 real-time images are checked to determine if they are of the same person. For example, a security parking lot may want to check if the person driving in is the same as the person driving out, 2 images are taken when the car enters and exists because it's unfeasible to build a database that stores every driver's photo for comparison. However, our experiment shows that pretrained Inception V1 net is not robust enough for this application. The self-compared distances increase unevenly, out of the 10 identities, the most significant increase is 52.38% and there is also one identity whose distance between himself decreases by 19.05%.

Despite the significant distance change, a threshold can still be set to conduct classification. The distance between one identity and himself is still much smaller than those between others. For example, in the "Glass to Glass" matrix, a threshold value of 0.4 will work perfectly to distinguish images belonging to the same person.

The other kind of application is when frontal image of the identity can be accessed in our database, in which case the change in distances between the "normal to normal" set and "normal to glass" set is of significance. For instance, the police usually have the frontal photo of the escaped criminal in their database, and they want to compare that with the photo taken by CCTV, which may be through the windshield of the car. In our dataset, the self-compared distance is increased by 114.29% on average, which means the insertion of glass do have a negative influence on the performance of our network simply because it makes the system "think" the difference between one identity is "bigger". However, in our classification applications, the relative magnitude of the value is of more significance, rather than absolute ones. A threshold of 0.6 also works perfectly in this case, because the distance between different identities is even bigger.

TABLE II. GLASS TO GLASS DISTANCE MATRIX

|  | 1_glass | 2_glass | 3_glass | 4_glass | 5_glass | 6_glass | 7_glass | 8_glass | 9_glass | 10_glass |
|---|---|---|---|---|---|---|---|---|---|---|
| **1_normal** | 0.42 | 1.25 | 1.03 | 1.13 | 1.28 | 1.06 | 1.19 | 1.01 | 1.27 | 1.32 |
| **2_normal** |  | 0.48 | 1.31 | 1.13 | 1.17 | 1.45 | 1.22 | 1.26 | 1.09 | 1.2 |
| **3_normal** |  |  | 0.52 | 1.17 | 1.1 | 0.86 | 0.85 | 0.96 | 1.2 | 1.24 |
| **4_normal** |  |  |  | 0.44 | 1.1 | 1.22 | 1.12 | 1.15 | 1.04 | 1.07 |
| **5_normal** |  |  |  |  | 0.38 | 1.21 | 1.03 | 1.3 | 0.93 | 0.94 |
| **6_normal** |  |  |  |  |  | 0.34 | 1 | 0.94 | 1.17 | 1.34 |
| **7_normal** |  |  |  |  |  |  | 0.4 | 1.1 | 1.17 | 1.2 |
| **8_normal** |  |  |  |  |  |  |  | 0.47 | 1.16 | 1.22 |
| **9_normal** |  |  |  |  |  |  |  |  | 0.51 | 0.99 |
| **10_normal** |  |  |  |  |  |  |  |  |  | 0.53 |

b. Threshold of 0.4 works if 2 images are of the same person

TABLE III. NORMAL TO GLASS DISTANCE MATRIX

|  | 1_glass | 2_glass | 3_glass | 4_glass | 5_glass | 6_glass | 7_glass | 8_glass | 9_glass | 10_glass |
|---|---|---|---|---|---|---|---|---|---|---|
| **1_glass** | 0.23 | 1.25 | 0.98 | 1.18 | 1.22 | 1.04 | 1.19 | 1.04 | 1.13 | 1.28 |
| **2_glass** |  | 0.26 | 1.2 | 1.19 | 1.14 | 1.42 | 1.2 | 1.33 | 1.1 | 1.16 |
| **3_glass** |  |  | 0.23 | 1.25 | 1.1 | 0.91 | 0.8 | 0.97 | 1.12 | 1.15 |
| **4_glass** |  |  |  | 0.26 | 1.09 | 1.23 | 1.19 | 1.18 | 1.04 | 1.1 |
| **5_glass** |  |  |  |  | 0.19 | 1.25 | 1.04 | 1.3 | 0.93 | 0.88 |
| **6_glass** |  |  |  |  |  | 0.17 | 0.98 | 0.91 | 1.06 | 1.29 |
| **7_glass** |  |  |  |  |  |  | 0.2 | 1.08 | 1.1 | 1.17 |
| **8_glass** |  |  |  |  |  |  |  | 0.32 | 1.11 | 1.19 |
| **9_glass** |  |  |  |  |  |  |  |  | 0.25 | 1 |
| **10_glass** |  |  |  |  |  |  |  |  |  | 0.32 |

c. Threshold of 0.4 works if 2 images are of the same person

*2) Image taken through water drop*

Similar to the glass factor, the model's performance in 2 kinds of applications is evaluated separately, "real-time" comparison and "database based" comparison where the frontal image is compared with the disturbed ones. By comparing the "normal to normal" distance matrix and "water to water" one, an average 167.86% increase of self-compared distance is observed. The increase is not uniform either, the highest increase is 331.25% and the lowest is 42.86%. Also, by comparing the "normal to water" and "normal to normal" matrix, an average increase of 285.55% between distance with one identity and himself is observed.

The relative magnitude of the biggest distance between any same identity and the smallest one between different people matters. In "water to water" distance matrix, though the difference "inside" one identity increases a lot, all of them are still smaller than the distance between different people, so threshold method can still be applied to easily finish the classification task. In our dataset, a threshold of 0.8 works.

However, it's not the case in the "normal to water" matrix, in which the system tries to classify based on a frontal normal image in database and a real-time disturbed photo. The "identity 7" has a self-compared distance of 1.05 and the distance between "6_normal" and "3_water" is just 0.99, the largest self-compared distance is smaller than the smallest distance between different identities, the threshold method for classification is failed.

TABLE IV. WATER TO WATER DISTANCE MATRIX

|  | 1_water | 2_water | 3_water | 4_water | 5_water | 6_water | 7_water | 8_water | 9_water | 10_water |
|---|---|---|---|---|---|---|---|---|---|---|
| **1_water** | 0.47 | 1.21 | 1.07 | 1.24 | 1.32 | 1.17 | 1.29 | 1.22 | 1.2 | 1.3 |
| **2_water** |  | 0.4 | 1.2 | 1.1 | 1.22 | 1.3 | 1.23 | 1.21 | 1.01 | 1.23 |
| **3_water** |  |  | 0.55 | 1.24 | 1.16 | 1.06 | 1.1 | 1.22 | 1.21 | 1.26 |
| **4_water** |  |  |  | 0.63 | 1.28 | 1.22 | 1.3 | 1.32 | 1.09 | 1.24 |
| **5_water** |  |  |  |  | 0.69 | 1.24 | 1.21 | 1.33 | 1.19 | 1.11 |
| **6_normal** |  |  |  |  |  | 0.39 | 1.13 | 1.21 | 1.21 | 1.32 |
| **7_normal** |  |  |  |  |  |  | 0.77 | 1.16 | 1.31 | 1.34 |
| **8_normal** |  |  |  |  |  |  |  | 0.53 | 1.25 | 1.29 |
| **9_normal** |  |  |  |  |  |  |  |  | 0.42 | 1.17 |
| **10_normal** |  |  |  |  |  |  |  |  |  | 0.57 |

d. Threshold of 0.8 works if 2 images are of the same person

TABLE V. NORMAL TO WATER DISTANCE MATRIX

|  | 1_normal | 2_normal | 3_normal | 4_normal | 5_normal | 6_normal | 7_normal | 8_normal | 9_normal | 10_normal |
|---|---|---|---|---|---|---|---|---|---|---|
| **1_water** | 0.64 | 1.31 | 1.06 | 1.2 | 1.31 | 1.14 | 1.21 | 1.13 | 1.3 | 1.35 |
| **2_water** |  | 0.8 | 1.25 | 1.04 | 1.22 | 1.3 | 1.15 | 1.12 | 1.1 | 1.23 |
| **3_water** |  |  | 0.79 | 1.19 | 1.17 | 0.99 | 0.91 | 1.09 | 1.26 | 1.31 |
| **4_water** |  |  |  | 0.8 | 1.21 | 1.21 | 1.25 | 1.19 | 1.16 | 1.21 |
| **5_water** |  |  |  |  | 0.84 | 1.2 | 1.05 | 1.3 | 1.17 | 1.18 |
| **6_normal** |  |  |  |  |  | 0.57 | 1 | 1.15 | 1.23 | 1.38 |
| **7_normal** |  |  |  |  |  |  | 1.05 | 1.23 | 1.38 | 1.43 |
| **8_normal** |  |  |  |  |  |  |  | 0.87 | 1.39 | 1.41 |
| **9_normal** |  |  |  |  |  |  |  |  | 0.73 | 1.13 |
| **10_normal** |  |  |  |  |  |  |  |  |  | 0.87 |

e. The system fails to classify if 2 images are of the same person

*3）Image taken through mist*

In places where the air humidity is high, it's common to see a layer of mist on glasses in the morning. The mist affects the quality of photos taken by CCTV greatly. The performance of the facial recognition model is tested using face images taken with this disturbing factor.

By comparing the "mist to mist" distances and "normal to normal" ones, an average increase of 134.01% in the self-compared distances is observed. In the "normal to mist" matrix, the distances "inside" each identity increases by 467.02%. The mist factor influences the recognition system by enlarging the difference between one identity and himself.

The "mist to mist" matrix is still feasible to classify based on a threshold value. In our dataset, 0.81 will perfectly distinguish the pairs belonging to the same person from the ones consists of different people's photos. That means in real application, if 2 images taken have a distance that is smaller than 0.81, then the system concludes that they belong to the same person, otherwise the system concludes that the images are of different identities.

However, that doesn't work in the Normal to Mist matrix for there is no such appropriate threshold value. The system does not work for "database based" applications which compares a photo taken with mist disturbing factors and the fontal one in our database.

TABLE VI.    MIST TO MIST DISTANCE MATRIX

|        | 1_mist | 2_mist | 3_mist | 4_mist | 5_mist | 6_mist | 7_mist | 8_mist | 9_mist | 10_mist |
|--------|--------|--------|--------|--------|--------|--------|--------|--------|--------|---------|
| 1_mist | 0.47   | 0.88   | 1.1    | 1.2    | 1.13   | 1.28   | 1.32   | 1.08   | 1.3    | 1.45    |
| 2_mist |        | 0.29   | 1.13   | 1.02   | 1.09   | 1.3    | 1.32   | 1.11   | 1.33   | 1.43    |
| 3_mist |        |        | 0.8    | 1.96   | 0.96   | 1.36   | 1.19   | 1.12   | 1.21   | 1.34    |
| 4_mist |        |        |        | 0.55   | 1.18   | 1.45   | 1.45   | 1.25   | 1.29   | 1.34    |
| 5_mist |        |        |        |        | 0.29   | 1.3    | 0.94   | 1.1    | 1.21   | 1.33    |
| 6_mist |        |        |        |        |        | 0.39   | 1.26   | 1.31   | 1.31   | 1.48    |
| 7_mist |        |        |        |        |        |        | 0.57   | 1.24   | 1.34   | 1.41    |
| 8_mist |        |        |        |        |        |        |        | 0.58   | 1.32   | 1.49    |
| 9_mist |        |        |        |        |        |        |        |        | 0.4    | 1.05    |
| 10_mist|        |        |        |        |        |        |        |        |        | 0.37    |

f. Threshold of 0.85 works if 2 images are of the same person

TABLE VII.    NORMAL TO MIST DISTANCE MATRIX

|         | 1_normal | 2_normal | 3_normal | 4_normal | 5_normal | 6_normal | 7_normal | 8_normal | 9_normal | 10_normal |
|---------|----------|----------|----------|----------|----------|----------|----------|----------|----------|-----------|
| 1_mist  | 1.32     | 1.36     | 1.43     | 1.47     | 1.46     | 1.29     | 1.46     | 1.42     | 1.36     | 1.42      |
| 2_mist  |          | 1.36     | 1.49     | 1.4      | 1.5      | 1.35     | 1.48     | 1.51     | 1.33     | 1.45      |
| 3_mist  |          |          | 1.35     | 1.51     | 1.34     | 1.31     | 1.36     | 1.4      | 1.29     | 1.39      |
| 4_mist  |          |          |          | 1.08     | 1.32     | 1.37     | 1.28     | 1.35     | 1.2      | 1.27      |
| 5_mist  |          |          |          |          | 1.29     | 1.16     | 1.23     | 1.25     | 1.27     | 1.38      |
| 6_mist  |          |          |          |          |          | 1.15     | 1.38     | 1.37     | 1.41     | 1.42      |
| 7_mist  |          |          |          |          |          |          | 1.26     | 1.23     | 1.42     | 1.47      |
| 8_mist  |          |          |          |          |          |          |          | 1.23     | 1.35     | 1.48      |
| 9_mist  |          |          |          |          |          |          |          |          | 0.72     | 1.13      |
| 10_mist |          |          |          |          |          |          |          |          |          | 0.84      |

g. The system fails to classify if 2 images are of the same person

## CONCLUSION

In this paper, a performance analysis of the human face recognition system is presented. This tested system consists of the pretrained MTCNN face detection network and Inception V1 face classification network. Facial images of ten identities with different disturbing factors were collected for experimental tests. Test data demonstrate that the system is able to perform classification by comparing real-time disrupted images. However, it fails to classify by comparing glass-factor image with normal frontal one in two datasets out of three. Further work can include testing on larger datasets or pre-train the model using glass-factor images rather than normal frontal ones.